\documentclass[11pt,a4paper]{article}

\usepackage[hyperref]{AACL-IJCNLP2020}
\usepackage{times}
\usepackage{latexsym}
\usepackage{url}

\usepackage{subfiles}
\usepackage{amssymb}
\usepackage{amsmath}

\usepackage{booktabs} 
\usepackage{xspace}
\usepackage{multirow}
\usepackage{tikz,tikz-qtree}
\usepackage{xfrac}
\usepackage{graphicx}
\usepackage{tablefootnote}
\usepackage{enumitem}

\aclfinalcopy 
\def\aclpaperid{424} 

\title{Contextualized End-to-End Neural Entity Linking}

\author{Haotian Chen \\
  BlackRock \\
  \small{\texttt{haotian.chen@blackrock.com}} \\ 
  \And
  Andrej Zukov-Gregoric \\
  BlackRock \\
  \small{\texttt{andrej.zukovgregoric@blackrock.com}}
  \AND
  Xi (David) Li \\
  BlackRock \\
  \small{\texttt{david.li@blackrock.com}}
 \And
  Sahil Wadhwa \\
  University of Illinois at Urbana-Champaign\thanks{{} {} Work done while at BlackRock.}\\
  \small{\texttt{sahilw2@illinois.edu}}
  }

\date{}

\begin{document}
\maketitle
\begin{abstract}
    We propose an entity linking (EL) model that jointly learns mention detection (MD) and entity disambiguation (ED). Our model applies task-specific heads on top of shared \texttt{BERT} contextualized embeddings. We achieve state-of-the-art results across a standard EL dataset using our model; we also study our model's performance under the setting when hand-crafted entity candidate sets are not available and find that the model performs well under such a setting also. 
\end{abstract}

\section{Introduction}
Entity linking (EL)\footnote{Also known as A2KB task in GERBIL evaluation platform \cite{gerbil} and end-to-end entity linking in some literature}, in our context, refers to the joint task of recognizing named entity mentions in text through mention detection (MD) and linking each mention to a unique entity in a knowledge base (KB) through entity disambiguation (ED)\footnote{Also known as D2KB task in GERBIL}. For example, in the sentence \textit{“The Times began publication under its current name in 1788,”} the span \textit{The Times} should be detected as a named entity mention and then linked to the corresponding entity: \href{https://en.wikipedia.org/wiki/The_Times}{\textit{The\_Times}}, a British newspaper. However, an EL model which disjointly applies MD and ED might easily mistake this mention with \href{https://en.wikipedia.org/wiki/The_New_York_Times}{\textit{The\_New\_York\_Times}}, an American newspaper. Since our model jointly learns MD and ED from the same contextualized \texttt{BERT} embeddings, its final EL prediction is partially informed by both. As a result, it is able to generalize better. 

Another common approach employed in previous EL research is candidate generation, where for each detected mention, a set of candidate entities is generated and the entities within it are ranked by a model to find the best match. Such sets are built using hand-crafted rules which define which entities make it in and which do not. This risks (1) skipping out on valid entities which should be in the candidate set and (2) inflating model performance since often times candidate sets contain only one or two items. These sets are almost always used at prediction time and sometimes even during training. Our model has the option of not relying on them during prediction, and never uses them during training. 

We introduce two main contributions:

\textit{(i)} We propose a new end-to-end differentiable neural EL model that jointly performs MD and ED and achieves state-of-the-art performance.

\textit{(ii)} We study the performance of our model when candidate sets are removed to see whether EL can perform well without them. 

\section{Related Work}
Neural-network based models have recently achieved strong results across standard EL datasets. Research has focused on learning better entity representations and extracting better local and global features through novel model architectures.

\noindent\textbf{Entity representation.} Good KB entity representations are a key component of most ED and EL models. Representation learning has been addressed by \citet{yamada-etal-2016-joint}, \citet{ganea-hofmann-2017-deep}, \citet{cao-etal-2017-bridge} and \citet{yamada2017learning}. \citet{sil2018neural} and \citet{cao-etal-2018-joint} extend it to the cross-lingual setting. More recently, \citet{yamada2019pre} have suggested learning entity representations using \texttt{BERT} which achieves state-of-the-art results in ED.

\noindent\textbf{Entity Disambiguation (ED).} The ED task assumes already-labelled mention spans which are then disambiguated. Recent work on ED has focused on extracting global features \citep{ratinov2011local, globerson2016collective, ganea-hofmann-2017-deep, le-titov-2018-improving}, extending the scope of ED to more non-standard datasets \citep{eshel-etal-2017-named}, and positing the problem in new ways such as building separate classifiers for KB entities \citep{barrena-etal-2018-learning}.

\noindent\textbf{Entity Linking (EL).} Early work by \citet{sil2013re}, \citet{luo2015joint} and \citet{nguyen2016j} introduced models that jointly learn NER and ED using engineered features. More recently, \citet{kolitsas-etal-2018-end} propose a neural model that first generates all combinations of spans as potential mentions and then learns similarity scores over their entity candidates. MD is handled implicitly by only considering mention spans which have non-empty candidate entity sets. \citet{martins-etal-2019-joint} propose training a multi-task NER and ED objective using a Stack-LSTM \citep{dyer-etal-2015-transition}. Finally, \citet{poerner2019bert} and \citet{broscheit-2019-investigating} both propose end-to-end EL models based on \texttt{BERT}. \citet{poerner2019bert} model the similarity between entity embeddings and contextualized word embeddings by mapping the former onto the latter whereas \citet{broscheit-2019-investigating} in essence do the opposite. Our work is different in three important ways: our training objective is different in that we explicitly model MD; we analyze the performance of our model when candidate sets are expanded to include the entire universe of entity embeddings; and we outperform both models by a wide margin.

\begin{figure*}[t]
\centering
\includegraphics[width=452pt,clip]{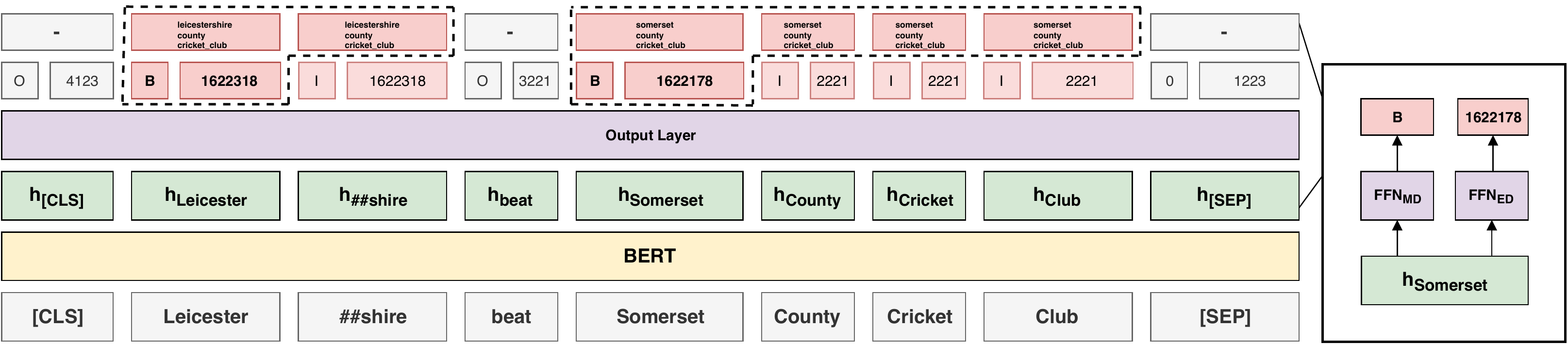}
\caption{Architecture of the proposed model. WordPiece tokens are passed through \texttt{BERT} forming contextualized embeddings. Each contextualized embedding is passed through two task-specific feed-forward neural networks for MD and ED, respectively. Entity ID prediction on the `B' MD tag is extended to the entire mention span. }
\label{fig:architecture}
\end{figure*}

\section{Model Description}
Given a document containing a sequence of $n$ tokens $\mathbf{w} = \{w_1, ..., w_n\}$ with mention label indicators\footnote{We use standard \textit{inside-outside-beginning} (IOB) tagging format introduced by \cite{ramshaw-marcus-1995-text}} $\mathbf{y}_{md} = \{I, O, B\}^n$ and entity IDs $\mathbf{y}_{ed} = \{j \in \mathbb{Z}: j \in [1, k]\}^n$ which index a pre-trained entity embedding matrix $\mathbf{E} \in \mathbb{R}^{k \times d}$ of entity universe size $k$ and entity embedding dimension $d$, the model is trained to tag each token with its correct mention indicator and link each mention with its correct entity ID.

\subsection{Text Encoder}
The text input to our model is encoded by \texttt{BERT} \cite{devlin-etal-2019-bert}. We initialize the pre-trained weights from \texttt{BERT-BASE}.\footnote{\href{https://github.com/google-research/bert}{https://github.com/google-research/bert}} The text input is tokenized by the cased WordPiece \cite{johnson-etal-2017-googles} sub-word tokenizer. The text encoder outputs $n$ contextualized WordPiece embeddings $\mathbf{h}$ which are grouped to form the embedding matrix $\mathbf{H} \in \mathbb{R}^{n \times m}$, where $m$ is the embedding dimension. In the case of \texttt{BERT-BASE}, $m$ is equal to $768$. 

The transformation from word level to WordPiece sub-word level labels is handled similarly to the \texttt{BERT} NER task, where the head WordPiece token represents the entire word, disregarding tail tokens.

\texttt{BERT} comes in two settings: feature-based and fine-tuned. Under the feature-based setting, \texttt{BERT} parameters are not trainable in the domain task (EL), whereas the fine-tuned setting allows \texttt{BERT} parameters to adapt to the domain task.

\subsection{EL model}
\noindent\textbf{MD} is modeled as a sequence labelling task. Contextualized embeddings $\mathbf{h}$ are passed through a feed-forward neural network and then softmaxed for classification over IOB:

\begin{align}
    \mathbf{m}_{md} &= \mathbf{W}_{md}\mathbf{h} + \mathbf{b}_{md}  \\
    \mathbf{p}_{md} &= \text{softmax}(\mathbf{m}_{md})  
\end{align}

\noindent where $\mathbf{b}_{md} \in \mathbb{R}^3$ is the bias term, $\mathbf{W}_{md} \in \mathbb{R}^{3\times m}$ is a weight matrix, and $\mathbf{p}_{md} \in \mathbb{R}^{3}$ is the predicted distribution across the $\{I,O,B\}$ tag set. The predicted tag is then simply:

\begin{equation}
    \mathbf{\hat{y}}_{md} = \underset{i}{\arg\max}\left\{\mathbf{p}_{md}(i)\right\} 
\end{equation}

\noindent\textbf{ED} is modeled by finding the entity (during inference this can be from either the entire entity universe or some candidate set) closest to the predicted entity embedding. We do this by applying an additional ED-specific feed-forward neural network to $\mathbf{h}$:

\begin{equation}
\begin{split}
    \mathbf{m}_{ed} &= \text{tanh}(\mathbf{W}_{ed}\mathbf{h} + \mathbf{b}_{ed})  \\
    \mathbf{p}_{ed} &= s(\mathbf{m}_{ed}, \mathbf{E}) \\ 
    \mathbf{\hat{y}}_{ed} &= \underset{j}{\arg\max} \left\{ \mathbf{p}_{ed}(j) \right\} 
\end{split}
\end{equation}

\noindent where $\mathbf{b}_{ed} \in \mathbb{R}^d$ is the bias term, $\mathbf{W}_{ed} \in \mathbb{R}^{d\times m}$ is a weight matrix, and $\mathbf{m}_{ed} \in \mathbb{R}^d$ is the same size as the entity embedding and $s$ is any similarity measure which relates $\mathbf{m}_{ed}$ to every entity embedding in $\mathbf{E}$. In our case, we use cosine similarity. Our predicted entity is the index of $\mathbf{p}_{ed}$ with the highest similarity score. 

We use pre-trained entity embeddings from \textit{wikipedia2vec} \cite{yamada2018wikipedia2vec}, as pre-training optimal entity representation is beyond the scope of this work. Ideally, pre-trained entity embeddings should be from a similar architecture to our EL model, but experiments show strong results even if they are not. The \textit{wikipedia2vec} entity embeddings used in our model are trained on the 2018 Wikipedia with $100$ dimensions and link graph support.\footnote{\href{https://wikipedia2vec.github.io/wikipedia2vec/pretrained/}{https://wikipedia2vec.github.io/wikipedia2vec/pretrained/}}

During inference, after receiving results for each token from both the MD and ED tasks, the mention spans are tagged with $\{B, I\}$ tags as shown in Figure \ref{fig:architecture}. For each mention span, the entity ID prediction of first token represents the entire mention span. The remaining non-mention and non-first entity ID prediction are masked out. Such behavior is facilitated by the training objective below.

During training, we minimize the following multi-task objective which is inspired by \citet{yolov2} from the object detection domain:\footnote{Similar to EL, object detection has two sub-tasks: locating bounding boxes and identifying objects in each box.}

\begin{equation}
 J(\theta) = \lambda\mathcal{L}_{md}(\theta) + (1-\lambda)\mathcal{L}_{ed}(\theta)
\end{equation}

\noindent where $\mathcal{L}_{md}$ is the cross entropy between predicted and actual distributions of IOB and $\mathcal{L}_{ed}$ is the cosine similarity between projected entity embeddings and actual entity embeddings. We tentatively explored triplet loss and contrastive loss with some simple negative mining strategies for ED but did not observe significant gains in performance. The two loss functions are weighted by a hyperparameter $\lambda$ (in our case $\lambda = 0.1$). Note that $\mathcal{L}_{md}$ is calculated for all non-pad head WordPiece tokens but $\mathcal{L}_{ed}$ is calculated only for the first WordPiece token of every labeled entity mention with a linkable and valid entity ID label. 

\section{Experiments}
\subsection{Dataset and Performance Metrics}

We train and evaluate our model on the widely used AIDA/CoNLL dataset \cite{hoffart2011robust}. It is a collection of news articles from Thomson Reuters, which is split into training, validation (testa) and test (testb) sets. Following convention, the evaluation metric is strong-matching span-level InKB micro and macro F1 score over gold mentions, where entity annotation is available \cite{gerbil}. Note that ED models are evaluated by accuracy metric while EL models are evaluated by F1, which penalizes the tagging of non-mention spans as entity mentions.

\subsection{Candidate Sets}\label{subsec:candidate_selection}
All EL models cited rely on candidate sets. As for our model, mentions can be efficiently disambiguated with respect to the entire entity universe, which we take to be the one million most frequent entities in 2018 Wikipedia. Consequently, our model can circumvent candidate generation, as well as the external knowledge that comes with it. In order to study the impact of candidate sets on our model, we apply candidate sets from \citet{hoffart2011robust} backed by the YAGO knowledge graph \cite{yago}. Importantly, we do not arbitrarily limit the size of the candidate sets. 

\subsection{Training Details and Settings}
 We train the EL model on the training split with a batch size of 4 for 50,000 steps. As in the original \texttt{BERT} paper, the model is optimized by the Adam optimizer \cite{kingma2014adam} with the same hyperparameters except the learning rate, which we set to be $\text{2e-5}$. Training was performed on a Tesla V100 GPU. A $0.1$ dropout rate was used on the prediction heads. Experiments are repeated three times to calculate an error range.

\subsection{Results}

\textbf{Comparison with Other EL Models.} We compare our model with six of the most recent, and best performing, EL models in Table \ref{el_results}. We study the performance of our model with, and without candidate sets (see Section \ref{subsec:candidate_selection}). We find that when candidate sets are provided, our model outperforms existing models by a significant margin. 

One of the problems of comparing results in the EL and ED space is that candidate sets are usually paper-specific and many works suggest their own methodologies for generating them. In addition to using candidate sets from \citet{hoffart2011robust} (which makes us comparable to \citet{kolitsas-etal-2018-end} who use the same sets), we impose no arbitrary limit on candidate set size. This means that many of our candidate sets have more than the standard 20-30 candidates, which are normally considered in past works.

Without candidate sets our model also shows good results and validation performance is on par with recent work by \citet{martins-etal-2019-joint} who used stack LSTMs \textit{with} candidate sets. We improve upon work by \citet{broscheit-2019-investigating} who, like us, do not use candidate sets. We use a larger overall entity universe (1M instead of 700K). Interestingly, \citet{broscheit-2019-investigating} note that during their error analysis only 3\% of wrong predictions were due to erroneous span detection. This could potentially explain our margin of improvement in the test set since our model is span-aware unlike theirs. For more details on the properties of the AIDA dataset we recommend \citet{ilievski-etal-2018-systematic}.

\noindent\textbf{Overfitting.} There are considerable drops in performance between validation and test both when BERT is fine-tuned or fixed, pointing to potential problems with overfitting. Identical behaviour is seen in \citet{broscheit-2019-investigating} and \citet{poerner2019bert}, who propose similar \texttt{BERT}-based models. Whether overfitting is due to BERT or the downstream models requires further research. 

Even more considerable drops in performance between validation and test are experienced when candidates sets are not used and entities are linked across the entire entity universe. We cannot be sure whether these drops are specific to BERT since no non-BERT works cite results over the entire entity universe.

\begin{table}[t]
\begin{centering}
\resizebox{\linewidth}{!}{%
\begin{tabular}{lcccc}
\hline 
 & \multicolumn{2}{c}{AIDA/testa F1 (val)} & \multicolumn{2}{c}{AIDA/testb F1 (test)}\tabularnewline
& Macro & Micro & Macro & Micro \\
\hline 
\citet{martins-etal-2019-joint} &  82.8 & 85.2 & 81.2  & 81.9 \tabularnewline
\citet{kolitsas-etal-2018-end} & 86.6 & 89.4 & 82.6 & 82.4 \tabularnewline
\citet{cao-etal-2018-joint} & 77.0 & 79.0 & 80.0 & 80.0 \tabularnewline
\citet{nguyen2016j} & - & - & - & 78.7 \tabularnewline
\citet{broscheit-2019-investigating} & - & 76.5 & - & 67.8 \tabularnewline
\citet{poerner2019bert} & 89.1 & 90.8 & 84.2 & 85.0 \tabularnewline

\hline
\hline
Fine-tuned BERT with candidate sets &  \textbf{92.6}$\pm0.2$ & \textbf{93.6}$\pm 0.2$ & \textbf{87.5}$\pm0.3$ & \textbf{87.7}$\pm 0.3$ \tabularnewline
Fine-tuned BERT without candidate sets &  82.6$\pm{0.2}$ & 83.5$\pm{0.2}$ & 70.7$\pm{0.3}$ & 69.4$\pm{0.3}$  \tabularnewline
\hline 
\end{tabular}}
\par\end{centering}
\centering{}
\caption{Strong-matching span-level InKB macro \& micro F1 results on validation and test splits of AIDA/CoNLL dataset. Note that the other models cited all use candidate sets. We run our models three times with different seeds to get bounds around our results.}
\label{el_results}
\end{table}

\noindent\textbf{Ablation Study.} We perform a simple ablation study, the results of which are shown in Table \ref{ablation_results}. We note that performance suffers in the EL task when \texttt{BERT} is not fine-tuned but still maintains strong results comparable to the state-of-the-art. Without fine-tuning, validation set performance decreases and becomes more comparable to test set performance, indicating that the fine-tuned \texttt{BERT} overfits in such a setting - we find this to be an interesting future direction of study.  

\noindent\textbf{Other Results.} Finally, during research, we swapped the Wikipedia2Vec entities with averaged-out 300-dimensional FastText embeddings \cite{bojanowski2017enriching} to see what the impact of not having entity-specific embeddings would be. To our surprise, the model performs on par with existing results which, we think, points to a combination of (1) \texttt{BERT} already having internal knowledge of entity-mentions given their context; and (2) many AIDA mentions being easily linkable by simply considering their surface-form. We think this too is an interesting direction of future study. Point (2) specifically points to the need for better EL datasets than AIDA, which was originally meant to be an ED dataset. A great study of point (1) can be found in \citet{poerner2019bert}.

\begin{table}
\begin{centering}\small
\resizebox{\linewidth}{!}{%
\begin{tabular}{lcccc}
\hline 
Ablation & \multicolumn{2}{c}{Validation F1} & \multicolumn{2}{c}{Test F1}\tabularnewline
& Macro & Micro & Macro & Micro \\
\hline 
Feature-based BERT with candidate sets &  87.1$\pm{0.1}$ & 90.3$\pm{0.1}$ & 83.5$\pm{0.3}$  & 84.8 $\pm{0.4}$ \tabularnewline
Feature-based BERT without candidate sets & 63.3$\pm1.1$ & 64.1$\pm 0.2$ & 57.2$\pm{0.2}$ & 54.1 $\pm0.3$ \tabularnewline
\hline
With fasttext entity embedding &  90.4 & 91.4& 82.8 & 82.9  \tabularnewline
\hline 
\end{tabular}}
\par\end{centering}
\centering{}
\caption{Ablation results on validation and test sets of AIDA/CoNLL. By feature-based BERT we mean BERT which is not fine-tuned to the task.}
\label{ablation_results}
\end{table}

\section{Conclusions and Future Work}
We propose an EL model that jointly learns the MD and ED task, achieving state-of-the-art results. We also show that training and inference without candidate sets is possible. We think that interesting future directions of study include a better understanding of how \texttt{BERT} already comprehends entities in text without reference to external entity embeddings. Finally, we think that moving forward, reducing the EL community's dependence on candidate sets could be a good thing and requires more research. Dropping candidate sets could make models more easily comparable.

\bibliography{main.bib}

\begin{thebibliography}{31}
\expandafter\ifx\csname natexlab\endcsname\relax\def\natexlab#1{#1}\fi

\bibitem[{Barrena et~al.(2018)Barrena, Soroa, and
  Agirre}]{barrena-etal-2018-learning}
Ander Barrena, Aitor Soroa, and Eneko Agirre. 2018.
\newblock \href {https://doi.org/10.18653/v1/K18-1017} {Learning text
  representations for 500{K} classification tasks on named entity
  disambiguation}.
\newblock In \emph{Proceedings of the 22nd Conference on Computational Natural
  Language Learning}, pages 171--180, Brussels, Belgium. Association for
  Computational Linguistics.

\bibitem[{Bojanowski et~al.(2017)Bojanowski, Grave, Joulin, and
  Mikolov}]{bojanowski2017enriching}
Piotr Bojanowski, Edouard Grave, Armand Joulin, and Tomas Mikolov. 2017.
\newblock Enriching word vectors with subword information.
\newblock \emph{Transactions of the Association for Computational Linguistics},
  5:135--146.

\bibitem[{Broscheit(2019)}]{broscheit-2019-investigating}
Samuel Broscheit. 2019.
\newblock \href {https://doi.org/10.18653/v1/K19-1063} {Investigating entity
  knowledge in {BERT} with simple neural end-to-end entity linking}.
\newblock In \emph{Proceedings of the 23rd Conference on Computational Natural
  Language Learning (CoNLL)}, pages 677--685, Hong Kong, China. Association for
  Computational Linguistics.

\bibitem[{Cao et~al.(2018)Cao, Hou, Li, Liu, Li, Chen, and
  Dong}]{cao-etal-2018-joint}
Yixin Cao, Lei Hou, Juanzi Li, Zhiyuan Liu, Chengjiang Li, Xu~Chen, and Tiansi
  Dong. 2018.
\newblock \href {https://doi.org/10.18653/v1/D18-1021} {Joint representation
  learning of cross-lingual words and entities via attentive distant
  supervision}.
\newblock In \emph{Proceedings of the 2018 Conference on Empirical Methods in
  Natural Language Processing}, pages 227--237, Brussels, Belgium. Association
  for Computational Linguistics.

\bibitem[{Cao et~al.(2017)Cao, Huang, Ji, Chen, and Li}]{cao-etal-2017-bridge}
Yixin Cao, Lifu Huang, Heng Ji, Xu~Chen, and Juanzi Li. 2017.
\newblock \href {https://doi.org/10.18653/v1/P17-1149} {Bridge text and
  knowledge by learning multi-prototype entity mention embedding}.
\newblock In \emph{Proceedings of the 55th Annual Meeting of the Association
  for Computational Linguistics (Volume 1: Long Papers)}, pages 1623--1633,
  Vancouver, Canada. Association for Computational Linguistics.

\bibitem[{Devlin et~al.(2019)Devlin, Chang, Lee, and
  Toutanova}]{devlin-etal-2019-bert}
Jacob Devlin, Ming-Wei Chang, Kenton Lee, and Kristina Toutanova. 2019.
\newblock \href {https://doi.org/10.18653/v1/N19-1423} {{BERT}: Pre-training of
  deep bidirectional transformers for language understanding}.
\newblock In \emph{Proceedings of the 2019 Conference of the North {A}merican
  Chapter of the Association for Computational Linguistics: Human Language
  Technologies, Volume 1 (Long and Short Papers)}, pages 4171--4186,
  Minneapolis, Minnesota. Association for Computational Linguistics.

\bibitem[{Dyer et~al.(2015)Dyer, Ballesteros, Ling, Matthews, and
  Smith}]{dyer-etal-2015-transition}
Chris Dyer, Miguel Ballesteros, Wang Ling, Austin Matthews, and Noah~A. Smith.
  2015.
\newblock \href {https://doi.org/10.3115/v1/P15-1033} {Transition-based
  dependency parsing with stack long short-term memory}.
\newblock In \emph{Proceedings of the 53rd Annual Meeting of the Association
  for Computational Linguistics and the 7th International Joint Conference on
  Natural Language Processing (Volume 1: Long Papers)}, pages 334--343,
  Beijing, China. Association for Computational Linguistics.

\bibitem[{Eshel et~al.(2017)Eshel, Cohen, Radinsky, Markovitch, Yamada, and
  Levy}]{eshel-etal-2017-named}
Yotam Eshel, Noam Cohen, Kira Radinsky, Shaul Markovitch, Ikuya Yamada, and
  Omer Levy. 2017.
\newblock \href {https://doi.org/10.18653/v1/K17-1008} {Named entity
  disambiguation for noisy text}.
\newblock In \emph{Proceedings of the 21st Conference on Computational Natural
  Language Learning ({C}o{NLL} 2017)}, pages 58--68, Vancouver, Canada.
  Association for Computational Linguistics.

\bibitem[{Ganea and Hofmann(2017)}]{ganea-hofmann-2017-deep}
Octavian-Eugen Ganea and Thomas Hofmann. 2017.
\newblock \href {https://doi.org/10.18653/v1/D17-1277} {Deep joint entity
  disambiguation with local neural attention}.
\newblock In \emph{Proceedings of the 2017 Conference on Empirical Methods in
  Natural Language Processing}, pages 2619--2629, Copenhagen, Denmark.
  Association for Computational Linguistics.

\bibitem[{Globerson et~al.(2016)Globerson, Lazic, Chakrabarti, Subramanya,
  Ringaard, and Pereira}]{globerson2016collective}
Amir Globerson, Nevena Lazic, Soumen Chakrabarti, Amarnag Subramanya, Michael
  Ringaard, and Fernando Pereira. 2016.
\newblock Collective entity resolution with multi-focal attention.
\newblock In \emph{Proceedings of the 54th Annual Meeting of the Association
  for Computational Linguistics (Volume 1: Long Papers)}, pages 621--631.

\bibitem[{Hoffart et~al.(2011)Hoffart, Yosef, Bordino, F{\"u}rstenau, Pinkal,
  Spaniol, Taneva, Thater, and Weikum}]{hoffart2011robust}
Johannes Hoffart, Mohamed~Amir Yosef, Ilaria Bordino, Hagen F{\"u}rstenau,
  Manfred Pinkal, Marc Spaniol, Bilyana Taneva, Stefan Thater, and Gerhard
  Weikum. 2011.
\newblock Robust disambiguation of named entities in text.
\newblock In \emph{Proceedings of the Conference on Empirical Methods in
  Natural Language Processing}, pages 782--792. Association for Computational
  Linguistics.

\bibitem[{Ilievski et~al.(2018)Ilievski, Vossen, and
  Schlobach}]{ilievski-etal-2018-systematic}
Filip Ilievski, Piek Vossen, and Stefan Schlobach. 2018.
\newblock \href {https://www.aclweb.org/anthology/C18-1056} {Systematic study
  of long tail phenomena in entity linking}.
\newblock In \emph{Proceedings of the 27th International Conference on
  Computational Linguistics}, pages 664--674, Santa Fe, New Mexico, USA.
  Association for Computational Linguistics.

\bibitem[{Johnson et~al.(2017)Johnson, Schuster, Le, Krikun, Wu, Chen, Thorat,
  Vi{\'e}gas, Wattenberg, Corrado, Hughes, and
  Dean}]{johnson-etal-2017-googles}
Melvin Johnson, Mike Schuster, Quoc~V. Le, Maxim Krikun, Yonghui Wu, Zhifeng
  Chen, Nikhil Thorat, Fernanda Vi{\'e}gas, Martin Wattenberg, Greg Corrado,
  Macduff Hughes, and Jeffrey Dean. 2017.
\newblock \href {https://doi.org/10.1162/tacl_a_00065} {{G}oogle{'}s
  multilingual neural machine translation system: Enabling zero-shot
  translation}.
\newblock \emph{Transactions of the Association for Computational Linguistics},
  5:339--351.

\bibitem[{Kingma and Ba(2014)}]{kingma2014adam}
Diederik~P Kingma and Jimmy Ba. 2014.
\newblock Adam: A method for stochastic optimization.
\newblock \emph{Proceedings of the 3rd International Conference on Learning
  Representations (ICLR)}.

\bibitem[{Kolitsas et~al.(2018)Kolitsas, Ganea, and
  Hofmann}]{kolitsas-etal-2018-end}
Nikolaos Kolitsas, Octavian-Eugen Ganea, and Thomas Hofmann. 2018.
\newblock \href {https://doi.org/10.18653/v1/K18-1050} {End-to-end neural
  entity linking}.
\newblock In \emph{Proceedings of the 22nd Conference on Computational Natural
  Language Learning}, pages 519--529, Brussels, Belgium. Association for
  Computational Linguistics.

\bibitem[{Le and Titov(2018)}]{le-titov-2018-improving}
Phong Le and Ivan Titov. 2018.
\newblock \href {https://doi.org/10.18653/v1/P18-1148} {Improving entity
  linking by modeling latent relations between mentions}.
\newblock In \emph{Proceedings of the 56th Annual Meeting of the Association
  for Computational Linguistics (Volume 1: Long Papers)}, pages 1595--1604,
  Melbourne, Australia. Association for Computational Linguistics.

\bibitem[{Luo et~al.(2015)Luo, Huang, Lin, and Nie}]{luo2015joint}
Gang Luo, Xiaojiang Huang, Chin-Yew Lin, and Zaiqing Nie. 2015.
\newblock Joint entity recognition and disambiguation.
\newblock In \emph{Proceedings of the 2015 Conference on Empirical Methods in
  Natural Language Processing}, pages 879--888.

\bibitem[{Martins et~al.(2019)Martins, Marinho, and
  Martins}]{martins-etal-2019-joint}
Pedro~Henrique Martins, Zita Marinho, and Andr{\'e} F.~T. Martins. 2019.
\newblock \href {https://doi.org/10.18653/v1/P19-2026} {Joint learning of named
  entity recognition and entity linking}.
\newblock In \emph{Proceedings of the 57th Annual Meeting of the Association
  for Computational Linguistics: Student Research Workshop}, pages 190--196,
  Florence, Italy. Association for Computational Linguistics.

\bibitem[{Nguyen et~al.(2016)Nguyen, Theobald, and Weikum}]{nguyen2016j}
Dat~Ba Nguyen, Martin Theobald, and Gerhard Weikum. 2016.
\newblock J-nerd: joint named entity recognition and disambiguation with rich
  linguistic features.
\newblock \emph{Transactions of the Association for Computational Linguistics},
  4:215--229.

\bibitem[{Poerner et~al.(2019)Poerner, Waltinger, and
  Sch{\"u}tze}]{poerner2019bert}
Nina Poerner, Ulli Waltinger, and Hinrich Sch{\"u}tze. 2019.
\newblock Bert is not a knowledge base (yet): Factual knowledge vs. name-based
  reasoning in unsupervised qa.
\newblock \emph{arXiv preprint arXiv:1911.03681}.

\bibitem[{Ramshaw and Marcus(1995)}]{ramshaw-marcus-1995-text}
Lance Ramshaw and Mitch Marcus. 1995.
\newblock \href {https://www.aclweb.org/anthology/W95-0107} {Text chunking
  using transformation-based learning}.
\newblock In \emph{Third Workshop on Very Large Corpora}.

\bibitem[{Ratinov et~al.(2011)Ratinov, Roth, Downey, and
  Anderson}]{ratinov2011local}
Lev Ratinov, Dan Roth, Doug Downey, and Mike Anderson. 2011.
\newblock Local and global algorithms for disambiguation to wikipedia.
\newblock In \emph{Proceedings of the 49th Annual Meeting of the Association
  for Computational Linguistics: Human Language Technologies-Volume 1}, pages
  1375--1384. Association for Computational Linguistics.

\bibitem[{{Redmon} and {Farhadi}(2017)}]{yolov2}
J.~{Redmon} and A.~{Farhadi}. 2017.
\newblock \href {https://doi.org/10.1109/CVPR.2017.690} {Yolo9000: Better,
  faster, stronger}.
\newblock In \emph{2017 IEEE Conference on Computer Vision and Pattern
  Recognition (CVPR)}, pages 6517--6525.

\bibitem[{R{\"{o}}der et~al.(2018)R{\"{o}}der, Usbeck, and Ngomo}]{gerbil}
Michael R{\"{o}}der, Ricardo Usbeck, and Axel{-}Cyrille~Ngonga Ngomo. 2018.
\newblock \href {https://doi.org/10.3233/SW-170286} {{GERBIL} - benchmarking
  named entity recognition and linking consistently}.
\newblock \emph{Semantic Web}, 9(5):605--625.

\bibitem[{Sil et~al.(2018)Sil, Kundu, Florian, and Hamza}]{sil2018neural}
Avirup Sil, Gourab Kundu, Radu Florian, and Wael Hamza. 2018.
\newblock Neural cross-lingual entity linking.
\newblock In \emph{Thirty-Second AAAI Conference on Artificial Intelligence}.

\bibitem[{Sil and Yates(2013)}]{sil2013re}
Avirup Sil and Alexander Yates. 2013.
\newblock Re-ranking for joint named-entity recognition and linking.
\newblock In \emph{Proceedings of the 22nd ACM international conference on
  Information \& Knowledge Management}, pages 2369--2374. ACM.

\bibitem[{Suchanek et~al.(2007)Suchanek, Kasneci, and Weikum}]{yago}
Fabian~M. Suchanek, Gjergji Kasneci, and Gerhard Weikum. 2007.
\newblock {Yago: A Core of Semantic Knowledge}.
\newblock In \emph{16th International Conference on the World Wide Web}, pages
  697--706.

\bibitem[{Yamada et~al.(2018)Yamada, Asai, Shindo, Takeda, and
  Takefuji}]{yamada2018wikipedia2vec}
Ikuya Yamada, Akari Asai, Hiroyuki Shindo, Hideaki Takeda, and Yoshiyasu
  Takefuji. 2018.
\newblock Wikipedia2vec: An optimized tool for learning embeddings of words and
  entities from wikipedia.
\newblock \emph{arXiv preprint 1812.06280}.

\bibitem[{Yamada and Shindo(2019)}]{yamada2019pre}
Ikuya Yamada and Hiroyuki Shindo. 2019.
\newblock Pre-training of deep contextualized embeddings of words and entities
  for named entity disambiguation.
\newblock \emph{arXiv preprint arXiv:1909.00426}.

\bibitem[{Yamada et~al.(2016)Yamada, Shindo, Takeda, and
  Takefuji}]{yamada-etal-2016-joint}
Ikuya Yamada, Hiroyuki Shindo, Hideaki Takeda, and Yoshiyasu Takefuji. 2016.
\newblock \href {https://doi.org/10.18653/v1/K16-1025} {Joint learning of the
  embedding of words and entities for named entity disambiguation}.
\newblock In \emph{Proceedings of The 20th {SIGNLL} Conference on Computational
  Natural Language Learning}, pages 250--259, Berlin, Germany. Association for
  Computational Linguistics.

\bibitem[{Yamada et~al.(2017)Yamada, Shindo, Takeda, and
  Takefuji}]{yamada2017learning}
Ikuya Yamada, Hiroyuki Shindo, Hideaki Takeda, and Yoshiyasu Takefuji. 2017.
\newblock Learning distributed representations of texts and entities from
  knowledge base.
\newblock \emph{Transactions of the Association for Computational Linguistics},
  5:397--411.

\end{thebibliography}
\bibliographystyle{acl_natbib}

\end{document}